\pdfoutput=1

\documentclass[a4paper,10pt,conference]{ieeeconf}

\IEEEoverridecommandlockouts

\usepackage{graphicx} 
\usepackage{xcolor} 
\usepackage{tabularx} 
\usepackage{makecell} 
\usepackage{balance} 
\usepackage[hyphens]{url} 

\newcolumntype{C}{>{\centering\arraybackslash}X}
\newcolumntype{L}{>{\raggedright\arraybackslash}X}

\title{\LARGE\bf
    OpenRoIS: A Community-Driven Open-Source Middleware\\
    Implementing the Robotic Interaction Service (RoIS) Framework\\
    for Physical Robots, Virtual Avatars, and Digital Services
}

\author{
    Sebastian Carrera Villalobos$^{1}$,
    Christopher Nolan Arellano$^{2}$,
    Arne Hitzmann$^{3}$,
    Edilson Morais Brito$^{3}$,\\
    Akira Utsumi$^{3}$,
    Yukiko Horikawa$^{3}$,
    Takahiro Miyashita$^{3}$,
    Hiroshi Ishiguro$^{3, 4}$,
    and Lotfi El Hafi$^{1, *}$
    \thanks{
        $^{1}$Sebastian Carrera Villalobos and Lotfi El Hafi are with Coarobo GK;
        2-2-2 Hikaridai, Seika, Soraku, Kyoto 619-0237, Japan. 
        {\tt\small \{sebastian, lotfi\}@coarobo.com}
    }
    \thanks{
        $^{2}$Christopher Nolan Arellano is with The University of Texas at Austin;
        110 Inner Campus Drive, Austin, Texas 78712, United States. 
        {\tt\small cna865@my.utexas.edu}
    }
    \thanks{
        $^{3}$Arne Hitzmann, Edilson Morais Brito, Akira Utsumi, Yukiko Horikawa, Takahiro Miyashita, and Hiroshi Ishiguro are with the Advanced Telecommunications Research Institute International~(ATR);
        2-2-2 Hikaridai, Seika, Soraku, Kyoto 619-0288, Japan. 
        {\tt\small \{arne.hitzmann, edilson, utsumi, horikawa, miyasita, ishiguro\}@atr.jp}
    }
    \thanks{
        $^{4}$Hiroshi Ishiguro is with The University of Osaka;
        1-3 Machikaneyama, Toyonaka, Osaka 560-8531, Japan. 
        \newline{\tt\small ishiguro@sys.es.osaka-u.ac.jp}
    }
    \thanks{
        $^{*}$Corresponding author.
    }
}

\begin{document}


\maketitle
\thispagestyle{empty}
\pagestyle{empty}


\begin{abstract}
    Service applications for human-robot interaction are commonly written against the hardware-specific interfaces of one platform, so a change of hardware forces a rewrite.
    The Robotic Interaction Service (RoIS) Framework 2.0, standardized by the Object Management Group (OMG), addresses this fragmentation with a platform-independent model in which Service Applications drive Human-Robot Interaction (HRI) Engines through standardized interfaces and symbolic messages.
    A specification alone, however, does not provide the maintained implementation, Software Development Kits (SDKs), and adapters that adoption requires.
    This paper presents OpenRoIS, a community-driven open-source middleware that implements RoIS 2.0, and takes the position that an openly developed, paradigm-neutral implementation is what carries the standard into practice.
    OpenRoIS contributes a recursive engine for both the main and the sub HRI Engines, a single protocol in which the RoIS method catalog over JSON-RPC 2.0 and WebSocket is the only interface at every hop, a single-source type pipeline for three languages, client SDKs for JavaScript, Python, and Unity, a component SDK, a browser inspector, and a Model Context Protocol (MCP) server for AI agents.
    These interfaces let a Service Application, scripted or an AI agent, address physical robots, virtual avatars, and digital services alike, each behind an adapter.
    In the functional validation, an AI agent and the browser inspector drove two physical robots from different vendors, a Reachy Mini and a Kachaka, behind one gateway, where a single command makes one robot greet a visitor while the other drives to the entrance.
    All of it is openly developed under the Apache-2.0 license at \url{https://openrois.org/}.
\end{abstract}


\begin{figure}[t]
    \centering
    \includegraphics[width=\linewidth]{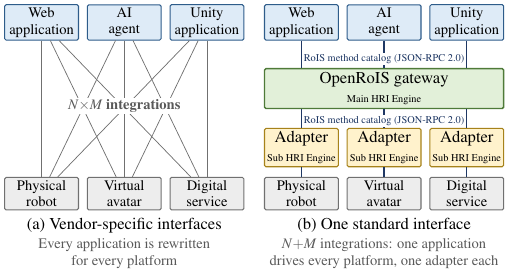}
    \caption{
        What OpenRoIS changes.
        (a) Without a standard interface, each of $N$ applications, such as a web application, an AI agent, or a Unity application, is integrated separately against each of $M$ platforms, whether physical robots, virtual avatars, or digital services.
        (b) OpenRoIS mediates the same systems through the symbolic interfaces of RoIS, so an application is written once and each platform integrated once by an adapter, with the same RoIS method catalog at every hop.
        Colors follow Fig.~\ref{fig:architecture} throughout.
    }
    \label{fig:concept}
\end{figure}


\begin{figure*}[t]
    \centering
    \includegraphics[width=\textwidth]{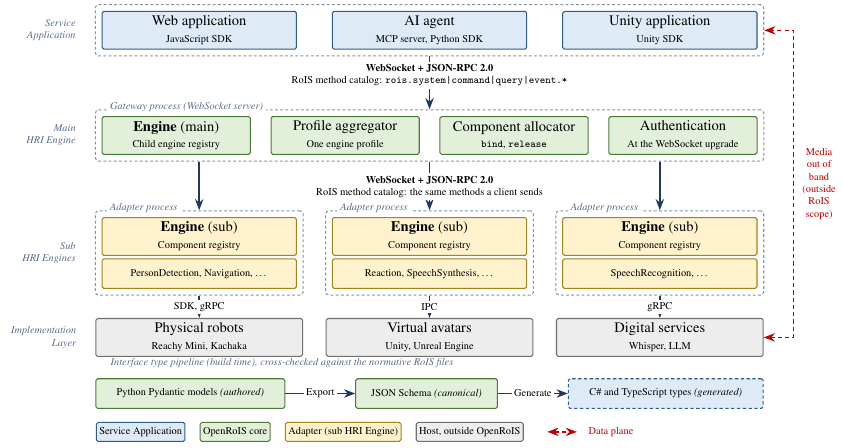}
    \caption{
        Architecture of OpenRoIS, with the RoIS concept realized by each row annotated on the left and the color legend at the bottom.
        The gateway is a control-plane router, each adapter chooses the backend transport its own host requires, and media stays on the data plane, which RoIS leaves out of scope.
        The hosts are named by kind with examples beneath, and the released adapters are those of the Reachy Mini and the Kachaka.
    }
    \label{fig:architecture}
\end{figure*}


\section{Introduction}
\label{sec:introduction}

Controlling physical robots from service applications has long suffered from a fragmentation problem, illustrated in Fig.~\ref{fig:concept}(a).
Each platform exposes its own hardware-specific interface, so any change of hardware forces an application rewrite, which undermines reusability and slows research transfer.
The cost grows multiplicatively rather than additively, since integrating $N$ applications against $M$ platforms yields $N \times M$ integrations built and maintained separately, none of which survives a change of platform.

The Robotic Interaction Service (RoIS) Framework~\cite{omg_rois_2026}, standardized by the Object Management Group (OMG), addresses this fragmentation by defining a platform-independent interaction model between Service Applications and Human-Robot Interaction (HRI) Engines.
RoIS 2.0 specifies hardware-independent HRI Components, hierarchical HRI Engines, profiles, and five interfaces, while leaving transport and implementation technologies open.
Its messages are symbolic, such as ``a person was detected'' or ``approach the person'', rather than raw sensor data and motor commands.
A specification alone, however, does not drive adoption.
Researchers and engineers need a maintained implementation with usable Software Development Kits (SDKs), reference adapters, and a network gateway.
To the best of our knowledge, no actively maintained open-source implementation of RoIS offers these.

This paper presents OpenRoIS as a concrete, openly developed implementation of RoIS 2.0, designed so that physical robots, virtual avatars, and digital services are driven through the same interfaces, whether by a scripted Service Application or by an AI agent, as shown in Fig.~\ref{fig:concept}(b).
Here a physical robot acts in a physical room, a virtual avatar is a character rendered by a 3D engine, a digital service is a bodiless capability such as speech recognition in the cloud, and an AI agent is an AI model that acts through tools in a harness.
Service Application keeps its RoIS meaning, the software that controls HRI Components through an HRI Engine, so an AI agent acting through OpenRoIS is one kind of Service Application.
OpenRoIS preserves the standard's platform-independent model while making explicit implementation choices, and it publishes source code, interface types, SDKs, adapters, and documentation under the Apache-2.0 license~\cite{openrois_2026}, with a public roadmap and contribution guide so that outside contributors have an entry point.
We take the position that an openly developed, paradigm-neutral implementation, and not a revision of the specification, is what carries RoIS into practice~\cite{quigley_ros_2009, macenski_robot_2022}.
The contributions of this paper are fourfold:
\begin{enumerate}
    \item A recursive engine architecture in which one engine class realizes both the main and the sub HRI Engines of the specification, so that a gateway and an adapter run the same engine, to any depth.
    \item A single protocol at every hop: the RoIS method catalog is the only interface between an application and the gateway, between a parent and a child HRI Engine, and between an engine and its HRI Components, so the middleware adds no interface of its own to the standard.
    \item A JSON-RPC 2.0 mapping of the RoIS interfaces over WebSocket, a single-source-of-truth type pipeline, client SDKs for JavaScript, Python, and Unity, a component SDK, and two ready-made applications, the OpenRoIS Inspector and an MCP server for AI agents.
    \item An open, community-driven release of the whole stack under the Apache-2.0 license, functionally validated end to end on two physical robots from different vendors.
\end{enumerate}


\section{Research Background}
\label{sec:background}

RoIS 2.0 defines a platform-independent model for interactions between Service Applications and HRI Engines~\cite{omg_rois_2026}.
It specifies hardware-independent HRI Components and hierarchical HRI Engines, shown on the left of Fig.~\ref{fig:architecture}, where sub HRI Engines encapsulate individual physical units and a main HRI Engine represents the overall system.
Service Applications interact through the main HRI Engine without depending on the sensors and actuators of the Implementation Layer or on the organization of sub HRI Engines.

The specification further defines the five interfaces summarized in Table~\ref{tab:interfaces}, 17 basic HRI Components, a profile mechanism for user-defined components, and, in version 2.0, streaming components for audio and video.
Command interaction follows a reservation lifecycle of \texttt{search}, \texttt{bind}, \texttt{execute}, \texttt{completed}, and \texttt{release}.

OpenRoIS complements rather than competes with existing robot middleware.
ROS~\cite{quigley_ros_2009} and ROS~2~\cite{macenski_robot_2022} are widely used in research robotics.
OpenRoIS can place ROS~2 behind an adapter, whose components reach ROS~2 services, actions, and topics through the optional ROS~2 support of the component SDK.
Service Applications therefore address physical robots on ROS~2 or other middleware, virtual avatars, and digital services through the same symbolic interface.
Open-RMF~\cite{open_robotics_rmf_2021} coordinates heterogeneous robot fleets and shared resources above individual robot control, and is complementary to RoIS.
RT-Middleware~\cite{ando_rt-middleware_2005} pioneered component-based robot middleware and led to the OMG Robotic Technology Component standard~\cite{omg_rtc_2012}, which RoIS references, while RSNP~\cite{narita_development_2009} defined a web-services protocol for networked robot services.
These efforts provide historical context for RoIS, while OpenRoIS contributes a maintained open-source implementation at the HRI service level.


\begin{table}[t]
    \caption{
        The five RoIS interfaces~\cite{omg_rois_2026} and their mapping onto the 16 JSON-RPC 2.0 methods and 4 notifications of OpenRoIS.
        Notifications are in italics, and the asterisk marks an OpenRoIS extension that the specification does not define.
    }
    \begin{center}
        \footnotesize
        \setlength{\tabcolsep}{3pt}
        \renewcommand{\arraystretch}{1.1}
        \begin{tabularx}{\columnwidth}{|l|L|l|}
            \hline
            \multicolumn{1}{|c|}{Interface} & \multicolumn{1}{c|}{Methods and notifications} & \multicolumn{1}{c|}{Namespace} \\
            \hline
            \hline
            System & \texttt{connect}, \texttt{disconnect}, \texttt{get\_profile}, \texttt{get\_error\_detail}, \textit{\texttt{notify\_error}}, \textit{\texttt{profile\_changed}}$^{*}$ & \texttt{rois.system.*} \\
            \hline
            Command & \texttt{search}, \texttt{bind}, \texttt{bind\_any}, \texttt{release}, \texttt{get/set\_parameter}, \texttt{execute}, \texttt{get\_command\_result}, \textit{\texttt{completed}} & \texttt{rois.command.*} \\
            \hline
            Query & \texttt{query} & \texttt{rois.query.*} \\
            \hline
            Event & \texttt{subscribe}, \texttt{unsubscribe}, \texttt{get\_event\_detail}, \textit{\texttt{notify\_event}} & \texttt{rois.event.*} \\
            \hline
            Streaming & None: streaming components are driven through the four interfaces above & --- \\
            \hline
        \end{tabularx}
    \end{center}
    \label{tab:interfaces}
\end{table}


\section{Scope and Non-Goals}
\label{sec:scope}

OpenRoIS manages the interaction plane between Service Applications and the physical robots, virtual avatars, and digital services they drive, that is, symbolic commands, queries, and events at the level of HRI Components.
In scope are the RoIS method catalog over JSON-RPC 2.0 and WebSocket, one engine class for gateways and adapters, a component SDK, three client SDKs, an MCP server, and the OpenRoIS Inspector.
Out of scope by design are high-frequency sensor logging, low-level real-time control loops, and fast time-series storage, which belong to native robot layers such as the Data Distribution Service~(DDS) of ROS~2 or to out-of-band telemetry, in keeping with the symbolic messages of RoIS.


\section{Proposed Architecture}
\label{sec:architecture}

\subsection{Design Principles}

Four implementation principles guide the design of OpenRoIS.
Paradigm neutrality keeps ROS, DDS, gRPC, and game engines out of the core engine and the client SDKs, so that such integrations reside in adapters.
Specification traceability requires every external interface and data type to trace back to the normative machine-readable artifacts of RoIS 2.0.
Transport separation gives the control plane a uniform carrier while each adapter selects the protocol its backend requires.
Client consistency means the same SDK calls address physical robots, virtual avatars, and digital services.

\subsection{Layered Architecture}

Fig.~\ref{fig:architecture} maps the OpenRoIS layers onto the corresponding RoIS concepts.
Service Applications use a client SDK to connect to the gateway, which realizes the main HRI Engine, while each adapter runs as a standalone sub HRI Engine for the components of one physical robot, virtual avatar, or digital service.
The gateway routes the control plane, aggregates the profiles of its child engines into one engine profile, and allocates components through binds, so that two clients cannot actuate the same component at once.
Media is designed to bypass the gateway, since a streaming component returns a descriptor of its stream and leaves the media transport to the deployment.
RoIS 2.0 standardizes stream control but leaves media encoding and transport technologies to implementations.

\subsection{Recursive Engine and Single Protocol}

OpenRoIS realizes the main/sub HRI Engine hierarchy of RoIS 2.0 through a recursive engine architecture, as shown in the two process compositions of Fig.~\ref{fig:architecture}.
A single \texttt{Engine} class manages local components and forwards RoIS calls to child engines.
The gateway combines this class with a server endpoint and a registry of child engines, whereas an adapter combines it with a client endpoint and a registry of local components.
The same dispatch implementation therefore realizes both the main and sub HRI Engine roles defined by RoIS 2.0, so that a gateway can also sit under another gateway.

An engine reaches its components with the same RoIS operations that a client sends to the engine, namely \texttt{execute}, \texttt{get\_command\_result}, \texttt{query}, \texttt{subscribe}, \texttt{unsubscribe}, \texttt{get\_event\_detail}, \texttt{get\_parameter}, \texttt{set\_parameter}, and \texttt{get\_error\_detail}.
A child engine is reached with these operations over a WebSocket, of which the parent engine is the only client, whereas a component in the same process is reached with the same operations as function calls on the same typed models.
Table~\ref{tab:hops} summarizes the resulting protocol at each hop: the RoIS method catalog is the only interface in the system, and the recursive engine and the single protocol are one idea, since an engine that speaks RoIS downward as well as upward can serve as gateway and as adapter to any depth.


\begin{table}[t]
    \caption{
        The protocol at each hop of OpenRoIS, where the RoIS method catalog is the only interface the middleware defines.
    }
    \begin{center}
        \footnotesize
        \setlength{\tabcolsep}{3pt}
        \renewcommand{\arraystretch}{1.1}
        \begin{tabularx}{\columnwidth}{|L|L|}
            \hline
            \multicolumn{1}{|c|}{Hop} & \multicolumn{1}{c|}{Protocol} \\
            \hline
            \hline
            Service Application to gateway (main HRI Engine) & RoIS method catalog, JSON-RPC 2.0 over WebSocket \\
            \hline
            Parent engine to child engine (gateway to adapter) & The same catalog over WebSocket, with the parent as the only client of the child \\
            \hline
            Engine to its own components & The same operations, in process \\
            \hline
            Component to its physical robot, virtual avatar, or digital service & Outside the middleware, the choice of the component (SDK, ROS~2, gRPC, HTTP) \\
            \hline
        \end{tabularx}
    \end{center}
    \label{tab:hops}
\end{table}


\subsection{Component Granularity}

A component is one RoIS HRI Component with one profile, that is, its commands, queries, events, and parameters, and an adapter groups the components of one physical robot, virtual avatar, or digital service.
A component author writes a Python class decorated with \texttt{@component(profile)}, whose methods are decorated with \texttt{@command}, \texttt{@query}, and \texttt{@event}, matching the three message lists of a RoIS HRI Component profile, as in Fig.~\ref{fig:listing}(a).
The engine supplies everything that the common messages of RoIS require, such as component status, parameter storage, start, stop, suspend, resume, binding, and identifiers, and parameters are applied through the standard \texttt{set\_parameter} command.
Applications compose the basic components where one fits and may declare a user-defined profile where none does, through the profile mechanism of RoIS, so that the standard component names guide but do not constrain the behaviors an application requires.


\begin{figure}[t]
    \centering
    \scriptsize
\begin{verbatim}
@component(NAVIGATION_PROFILE)
class Navigation(Component):
    @command("start")
    async def start(self) -> dict[str, object]:
        target = self.parameters["target_positions"][0]
        await self._robot.go_to(target)
        self.emit("reached_target", target=target,
                  is_final_target=True)
        return {}
    @event("reached_target")
    async def reached_target(self) -> None:
        pass
\end{verbatim}
    \vspace{-1.5ex}{\footnotesize (a)}\vspace{-0.5ex}
\begin{verbatim}
GATEWAY = "ws://localhost:8765"
REFS = ["reachy_mini/person_detection",
        "kachaka/person_detection"]
async with RoISClient.connect(GATEWAY) as client:
    client.on("rois.event.notify_event", print)
    for ref in REFS:
        await client.bind(ref)
        await client.subscribe("person_detected",
                               component_ref(ref))
        await client.execute([{"component_ref": ref,
                               "command_type": "start"}])
    # Both robots now report the persons they detect.
\end{verbatim}
    \vspace{-1.5ex}{\footnotesize (b)}
    \caption{
        (a) A Navigation component written with the component SDK, abridged from its documentation, where the decorators bind Python methods to the messages of the RoIS profile.
        (b) A Service Application written with the Python SDK, where the prefix of a component reference names the sub HRI Engine of a robot and the same calls serve both robots.
    }
    \label{fig:listing}
\end{figure}


\section{Middleware Implementation}
\label{sec:implementation}

\subsection{Wire Protocol}

In OpenRoIS, the System, Command, Query, and Event interfaces of RoIS map to 16 JSON-RPC 2.0 methods and 4 notifications in the namespaces of Table~\ref{tab:interfaces}, carried over WebSocket.
Requests receive responses, whereas command completions, events, errors, and profile changes are delivered as JSON-RPC notifications, preserving the asynchronous interaction semantics of RoIS 2.0.
Carrying the catalog over WebSocket is a concrete OpenRoIS transport choice permitted by the transport-neutral specification and supports browser-based operation.
OpenRoIS documents six extensions that the specification does not define, among them the profile-change notification and the event payload carried in the event notification, which saves one \texttt{get\_event\_detail} call.
The Streaming interface has no methods of its own, since a streaming component is driven through the other four interfaces.
The gateway listens on the loopback interface by default, and remote access goes through an authenticated reverse proxy or a virtual private network that also provides encryption.

\subsection{Interface Type Pipeline}

OpenRoIS maintains Python data models corresponding to the RoIS message types, exports them to a shared JSON Schema representation, and generates C\# and TypeScript types, as shown in Fig.~\ref{fig:architecture}.
This single-source pipeline prevents type drift between the engine and the SDKs.
It covers all 18 normative profiles of RoIS 2.0 with an XML profile codec, and its tests number 1,211 in Python, 661 in TypeScript, and 935 in C\#, with about 300 more that cross-check the generated models against the normative machine-readable files of the specification when those files are supplied.

\subsection{SDKs, Components, and Applications}

OpenRoIS provides three client SDKs that expose the RoIS interfaces to Service Applications, a JavaScript SDK for browsers and Node.js, a Python SDK, and a Unity SDK.
A component SDK in Python builds sub HRI Engines from components, with shared resources and optional ROS~2 support.
The eight packages, namely the interfaces, engine, component SDK, gateway, three client SDKs, and MCP server, are released together on PyPI and npm, with the Unity SDK installed by Git URL, and a single command, \texttt{docker compose up --build}, starts a gateway, a simulated adapter, and the Inspector for a browser.
Reference components run on the Reachy Mini\footnote{\url{https://www.pollen-robotics.com/}}, an open-source physical robot for the desktop by Pollen Robotics, whose adapter provides the SystemInformation, PersonDetection, SpeechSynthesis, and Reaction components on the physical robot and in its simulator, while a simulated adapter also provides Navigation.
A second adapter serves the Kachaka, a shelf carrier by Preferred Robotics, with the SystemInformation, PersonDetection, Navigation, and Move components over the gRPC API of the robot\footnote{Both adapters are at \url{https://github.com/openrois}.}.
The same client code drives either robot, as in Fig.~\ref{fig:listing}(b), with only the component reference changing.

Two ready-made Service Applications build on the SDKs.
The OpenRoIS Inspector, shown in Fig.~\ref{fig:inspector}, is a browser application that displays the engines and components behind a gateway, binds components, runs their commands and command sequences with concurrent branches, follows their events, and records the protocol traffic, with every form built from the component profiles.
The MCP server\footnote{\url{https://modelcontextprotocol.io/}} lets an AI agent discover components, read their profiles, follow events, and run commands without knowing the interface of any particular physical robot or virtual avatar.
The server is read-only by default, and its tools that bind, release, set parameters, and run commands appear only when the operator allows actions, optionally for named components only.


\newcommand{\inspectorshot}[2]{%
    \setlength{\unitlength}{\linewidth}%
    \frame{\begin{picture}(1,0.5622)%
        \put(0,0){\includegraphics[width=\linewidth]{#1}}%
        \put(0.985,0.012){\makebox(0,0)[rb]{\colorbox{white}{\footnotesize #2}}}%
    \end{picture}}%
}
\begin{figure}[!t]
    \centering
    \inspectorshot{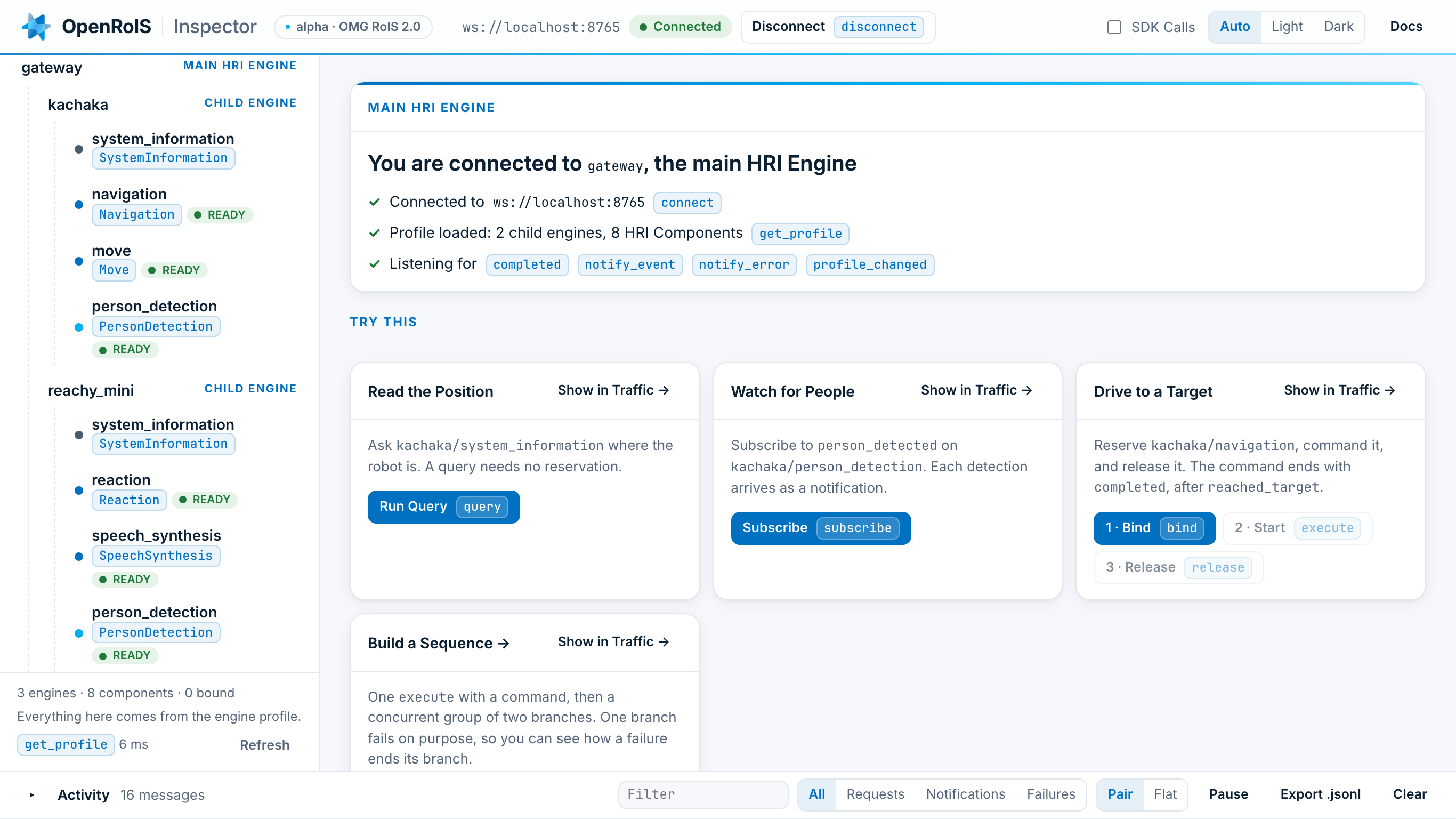}{(a)}\\[5pt]
    \inspectorshot{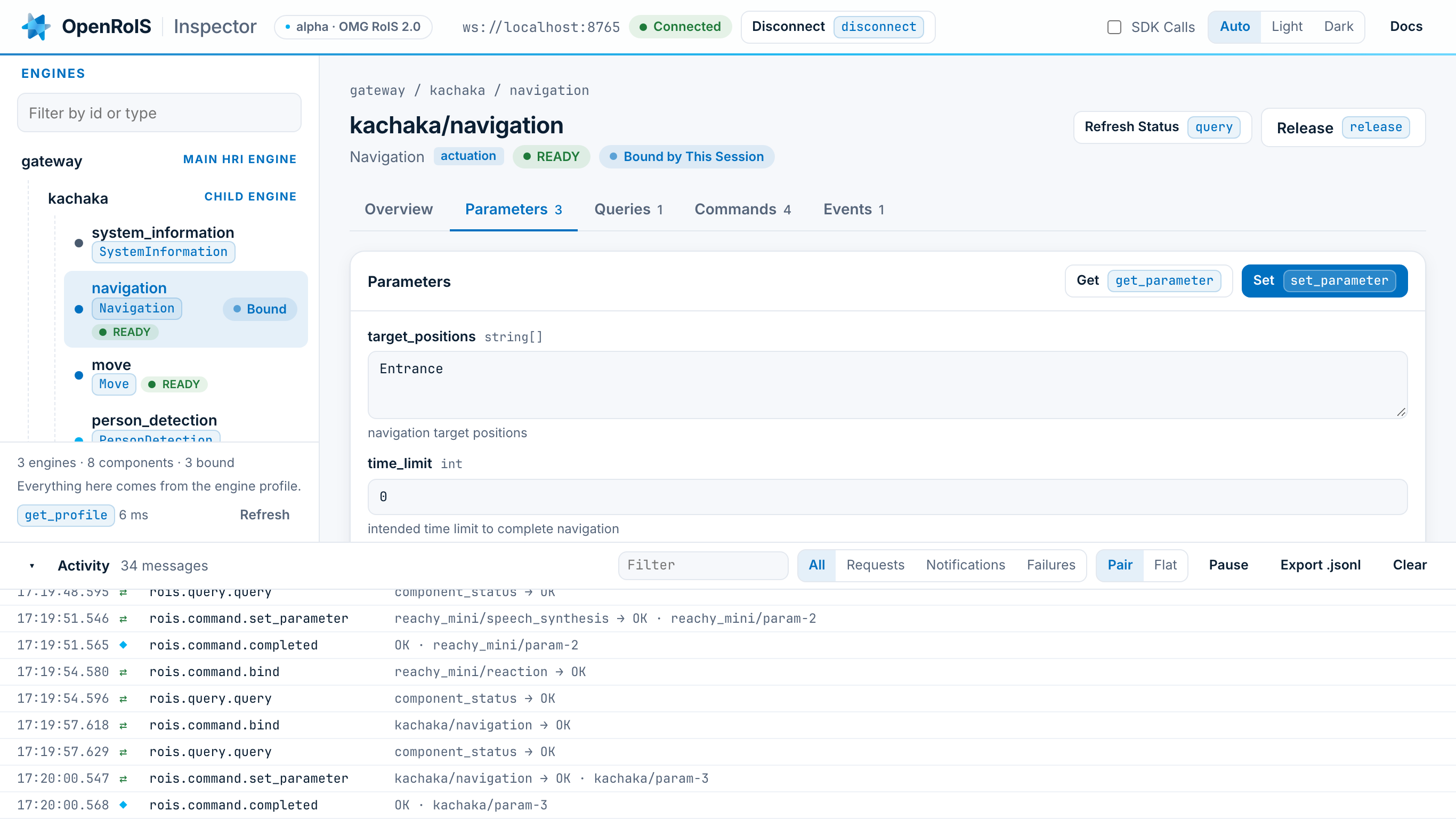}{(b)}\\[5pt]
    \inspectorshot{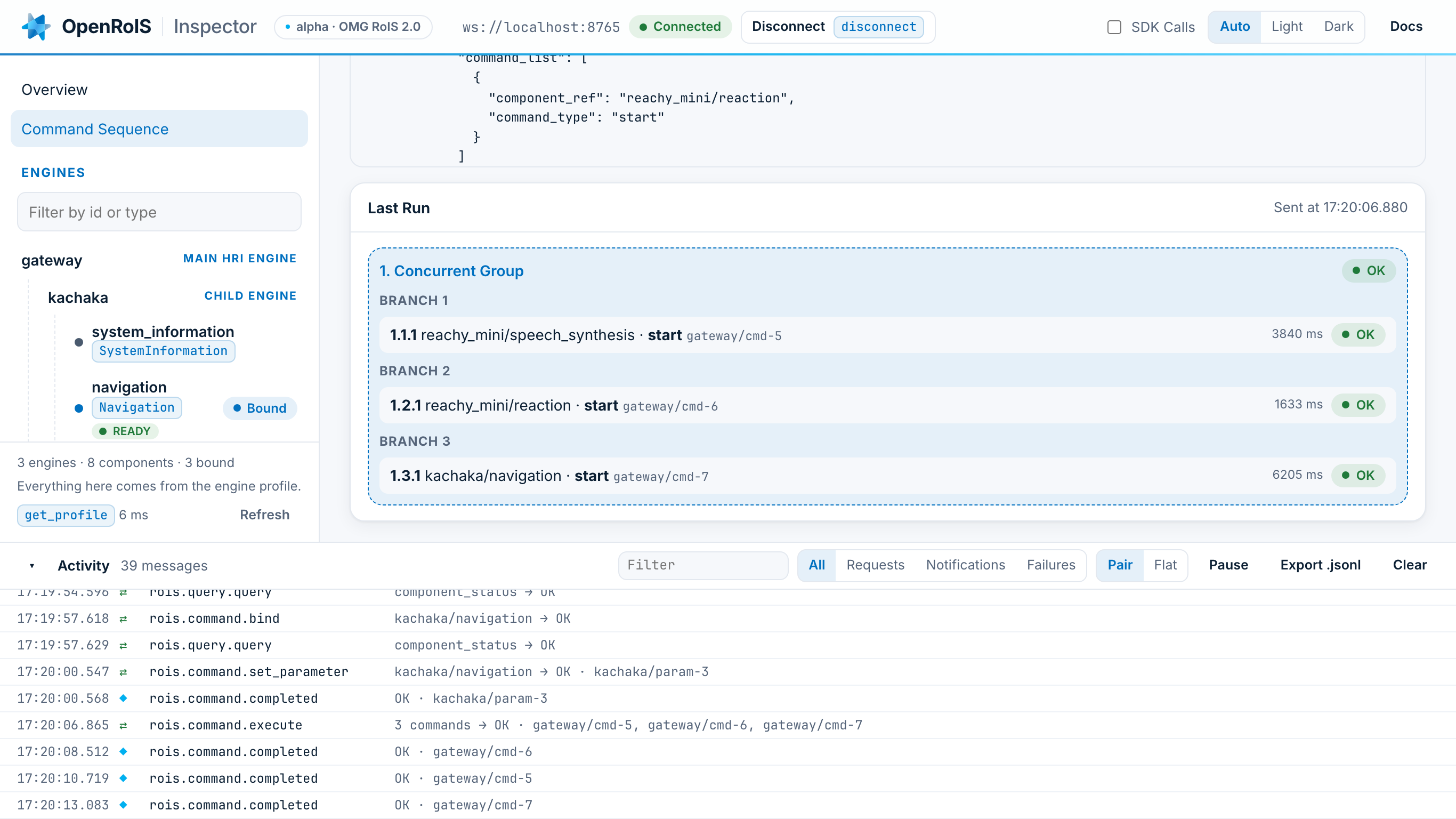}{(c)}
    \caption{
        The OpenRoIS Inspector connected to a gateway that serves both adapters, here with both robots simulated.
        (a) The main HRI Engine with its two sub HRI Engines and their eight components.
        (b) The Navigation component of the Kachaka, bound, with the entrance set as its target and the protocol traffic at the bottom.
        (c) One \texttt{execute} whose three concurrent branches make the Reachy Mini speak and nod while the Kachaka drives, run to its end.
    }
    \label{fig:inspector}
\end{figure}


\section{Functional Validation}
\label{sec:validation}

\subsection{End to End on Hardware}

OpenRoIS was validated functionally rather than benchmarked, in a scenario where one application greets a visitor with the Reachy Mini and sends the Kachaka to the entrance.
Once the Reachy Mini detects the visitor, a single command with three concurrent branches makes it speak and nod while the Kachaka sets off.
As Table~\ref{tab:validation} lists, an AI agent ran this scenario on the two physical robots through the MCP server from a single request, the Inspector drove both, as in Fig.~\ref{fig:inspector}, and the 11-step script of the Reachy Mini adapter also covered an error of the daemon of that robot and the recovery of every component.
A physical and a simulated Reachy Mini also ran behind one gateway under one application.
This validation involved no human participants, so no review by an institutional review board was required.


\begin{table}[t]
    \caption{
        The functional validation: three Service Applications drove two physical robots through the same interfaces ($\bullet$:~run on the physical robot).
    }
    \begin{center}
        \footnotesize
        \setlength{\tabcolsep}{3pt}
        \renewcommand{\arraystretch}{1.1}
        \begin{tabularx}{\columnwidth}{|L|c|c|}
            \hline
            \multicolumn{1}{|c|}{Service Application} & Reachy Mini & Kachaka \\
            \hline
            \hline
            Scripted application, Python SDK & $\bullet$ & $\bullet$ \\
            \hline
            OpenRoIS Inspector, JavaScript SDK & $\bullet$ & $\bullet$ \\
            \hline
            AI agent, MCP server & $\bullet$ & $\bullet$ \\
            \hline
            \hline
            Behind the adapter & Python SDK & gRPC API \\
            \hline
        \end{tabularx}
    \end{center}
    \label{tab:validation}
\end{table}


\subsection{Wire Scenarios and Tests}

One list of 24 scenarios specifies the contract between a gateway and its clients: each of the three client SDKs passes its 21 client scenarios, 19 of the 24 also pass on raw JSON-RPC with no SDK, and each SDK is tested against the published interface packages before its own release.
At the time of writing, the repository holds about 3,700 automated tests, and continuous integration runs up to 23 checks, each on the changes that concern it.
Implementing the specification closely also surfaced ambiguities and gaps, which the repository records, such as the \texttt{reached\_target} event of the Navigation component, present in its machine-readable profile but missing from the corresponding table of the specification.

\subsection{Scalability and Fault Tolerance}

The centralized gateway simplifies discovery and routing but concentrates the control plane in one process.
One gateway process serves several clients and several adapters at once, binds keep two clients from actuating the same component, adapters reconnect with a growing delay, and the gateway can restart without its adapters refusing its commands.
Gateway redundancy, automatic reconnection of the clients in the SDKs, load testing, and latency and throughput measurements are left to future work.


\section{Field Application}
\label{sec:application}

The practical case for a standard interface appears wherever several groups deploy physical robots side by side.
At Avatar Land, a 19-day public event in Osaka, eleven distinct demonstrations ran together, each integrated separately~\cite{el_hafi_public_2025}.
One of them drove three physical robots, a Fetch Robotics mobile manipulator and two Preferred Robotics Kachaka shelf carriers, together with a Meta Quest 3 headset as a single autonomous system, wired together by hand over a common middleware~\cite{quigley_ros_2009, el_hafi_abstraction-rich_2018, el_hafi_software_2022}.
Symbolically that scenario is what RoIS describes, since it detects a person, identifies the object indicated, and navigates to fetch it, yet none of its implementation logic was portable to the other demonstrations.
With OpenRoIS, such service logic can be written once against the RoIS interfaces and reused across embodiments, each integrated by its own adapter.

Because the OpenRoIS control plane is uniform, deployments of very different scales differ mainly in where the processes run.
The processes may share one machine, which is the development case.
A gateway on a local network may instead serve several physical robots, each with its own adapter, while presenting one aggregated profile to the application.
Across the internet, the operator application, gateway, and each host may run in three different places, as in a teleoperation deployment.
Components needing more compute than a physical robot carries, such as speech recognition, can run as digital services in a separate adapter without exposing their location to the application.

OpenRoIS serves two kinds of embodiment, and AI agents drive both.
Physical robots are the platforms at hand, on which every layer was validated, from the interfaces to the applications.
The same architecture also supports virtual avatars, characters rendered by a 3D engine and driven by an operator or an autonomous policy, each as one more adapter, and the Unity SDK connects Unity to the same gateway.
AI agents already sit above both through the MCP server, which lets an AI agent that reasons in symbols rather than joint angles bind whichever physical robot or virtual avatar is available.

The avatars of the JST Moonshot R\&D Program, which Avatar Land demonstrated in public, are where the two kinds meet.
The avatar-symbiotic society that the program pursues lets humans act remotely through robotic and virtual embodiments~\cite{ishiguro_realisation_2021}, and such services stress every property OpenRoIS is designed around.
OpenRoIS was developed independently of that program.
An application must reach an embodiment across the internet, the embodiment may be a physical robot one day and a virtual avatar the next, and live audio and video must flow alongside symbolic control.


\section{Conclusion}
\label{sec:conclusion}

This paper presented OpenRoIS, a community-driven open-source middleware providing a concrete implementation of the OMG RoIS Framework 2.0 for physical robots, virtual avatars, and digital services, driven by scripted applications and AI agents alike.
OpenRoIS realizes the roles and interfaces defined by the standard through a recursive engine architecture, a single protocol at every hop, a JSON-RPC 2.0 mapping over WebSocket, a single-source-of-truth type pipeline, three client SDKs, a component SDK, and two ready-made applications.
The work separates the platform-independent model defined by RoIS 2.0 from the concrete implementation choices required to put it into practice.

The stack has been validated end to end with two physical robots from different vendors and three Service Applications, with no interface added to the standard, while the OpenRoIS APIs remain subject to change until a stable release.
The public roadmap, which carries no dates, plans RoIS conformance, recovery when a gateway or an adapter restarts, a physical robot and a virtual avatar under one application, authentication and authorization, media streaming, and a stable API, and later the remaining basic HRI Components, a registry of community components and adapters, and a Hub for operators.
Community-driven here means that OpenRoIS is built in the open.
An adapter for a new physical robot, in its own repository, is the natural entry point, and contributions are welcome at \url{https://github.com/openrois}.


\section*{Acknowledgment}

The authors thank the Avatar-Symbiotic Society~\cite{ishiguro_realisation_2021} project of the JST Moonshot R\&D Program for the cooperation that enabled the first trials of OpenRoIS at ATR.
The authors also thank Bill Fagelson of The University of Texas at Austin for his participation in the early discussions.


\balance

\bibliographystyle{IEEEtran}
\bibliography{references}

@article{ishiguro_realisation_2021,
	title = {The {Realisation} of an {Avatar}-{Symbiotic} {Society} {Where} {Everyone} can {Perform} {Active} {Roles} without {Constraint}},
	volume = {35},
	issn = {0169-1864},
	url = {https://doi.org/10.1080/01691864.2021.1928548},
	doi = {10.1080/01691864.2021.1928548},
	number = {11},
	journal = {Advanced Robotics (AR)},
	author = {Ishiguro, Hiroshi},
	month = jun,
	year = {2021},
	pages = {650--656},
}

@inproceedings{quigley_ros_2009,
	address = {Kobe, Japan},
	title = {{ROS}: {An} {Open}-{Source} {Robot} {Operating} {System}},
	url = {http://www.willowgarage.com/papers/ros-open-source-robot-operating-system},
	booktitle = {2009 {IEEE} {Workshop} on {Open} {Source} {Software}},
	author = {Quigley, Morgan and Gerkey, Brian and Conley, Ken and Faust, Josh and Foote, Tully and Leibs, Jeremy and Berger, Eric and Wheeler, Rob and Ng, Andrew},
	month = may,
	year = {2009},
	pages = {1--6},
}

@article{el_hafi_software_2022,
	title = {Software {Development} {Environment} for {Collaborative} {Research} {Workflow} in {Robotic} {System} {Integration}},
	volume = {36},
	issn = {0169-1864},
	url = {https://doi.org/10.1080/01691864.2022.2068353},
	doi = {10.1080/01691864.2022.2068353},
	number = {11},
	journal = {Advanced Robotics (AR)},
	author = {El Hafi, Lotfi and Garcia Ricardez, Gustavo Alfonso and von Drigalski, Felix and Inoue, Yuki and Yamamoto, Masaki and Yamamoto, Takashi},
	month = jun,
	year = {2022},
	pages = {533--547},
}

@misc{omg_rois_2026,
	title = {Robotic {Interaction} {Service} ({RoIS}) {Framework}, {Version} 2.0},
	url = {https://www.omg.org/spec/RoIS/2.0},
	author = {{Object Management Group}},
	month = jun,
	year = {2026},
	note = {{OMG} {Document} formal/26-06-03},
}

@article{macenski_robot_2022,
	title = {Robot {Operating} {System} 2: {Design}, {Architecture}, and {Uses} in the {Wild}},
	volume = {7},
	number = {66},
	pages = {eabm6074},
	journal = {Science Robotics},
	author = {Macenski, Steven and Foote, Tully and Gerkey, Brian and Lalancette, Chris and Woodall, William},
	month = may,
	year = {2022},
}

@inproceedings{ando_rt-middleware_2005,
	address = {Edmonton, Canada},
	title = {{RT}-{Middleware}: {Distributed} {Component} {Middleware} for {RT} ({Robot} {Technology})},
	booktitle = {2005 {IEEE}/{RSJ} {International} {Conference} on {Intelligent} {Robots} and {Systems} ({IROS} 2005)},
	author = {Ando, Noriaki and Suehiro, Takashi and Kitagaki, Kosei and Kotoku, Tetsuo and Yoon, Woo-Keun},
	month = aug,
	year = {2005},
	pages = {3555--3560},
}

@article{narita_development_2009,
	title = {Development of {RSNP} ({Robot} {Service} {Network} {Protocol}) 2.0 {Targeting} a {Robot} {Service} {Platform} in {Diffusion} {Period}},
	volume = {27},
	number = {8},
	pages = {857--867},
	journal = {Journal of the Robotics Society of Japan},
	author = {Narita, Masahiko and Murakawa, Yoshihiko and Ueki, Miwa and others},
	year = {2009},
	note = {(in Japanese)},
}

@misc{omg_rtc_2012,
	title = {Robotic {Technology} {Component} ({RTC}), {Version} 1.1},
	url = {https://www.omg.org/spec/RTC},
	author = {{Object Management Group}},
	year = {2012},
}

@inproceedings{el_hafi_public_2025,
	address = {Osaka, Japan},
	title = {Public {Evaluation} on {Potential} {Social} {Impacts} of {Fully} {Autonomous} {Cybernetic} {Avatars} for {Physical} {Support} in {Daily}-{Life} {Environments}: {Large}-{Scale} {Demonstration} and {Survey} at {Avatar} {Land}},
	booktitle = {2025 {IEEE} {International} {Conference} on {Advanced} {Robotics} and its {Social} {Impacts} ({ARSO} 2025)},
	author = {El Hafi, Lotfi and Onishi, Kazuma and Hasegawa, Shoichi and others},
	year = {2025},
}

@misc{open_robotics_rmf_2021,
	title = {Programming {Multiple} {Robots} with {ROS}~2},
	url = {https://osrf.github.io/ros2multirobotbook/},
	author = {{Open Robotics}},
	note = {{Open}-{RMF} documentation. Accessed: Aug. 28, 2026},
}

@misc{openrois_2026,
	title = {{OpenRoIS}: {Open}-{Source} {Middleware} for the {OMG} {RoIS} {Framework} 2.0},
	url = {https://github.com/openrois/openrois},
	author = {{OpenRoIS Community}},
	year = {2026},
	note = {{Apache-2.0} open-source software},
}

@inproceedings{el_hafi_abstraction-rich_2018,
	address = {Kasugai, Japan},
	title = {Abstraction-rich {Workflow} for {Agile} {Collaborative} {Development} and {Deployment} of {Robotic} {Solutions}},
	url = {https://www.rsj.or.jp/data_acrsj/rom_data/g36-rom/pdf/3D3-02.pdf},
	booktitle = {2018 {Annual} {Conference} of the {Robotics} {Society} of {Japan} ({RSJ} 2018)},
	author = {El Hafi, Lotfi and Hagiwara, Yoshinobu and Taniguchi, Tadahiro},
	month = sep,
	year = {2018},
	pages = {1--3},
}


\end{document}